# IK-PSO, PSO Inverse Kinematics Solver with Application to Biped Gait Generation

Nizar Rokbani, Adel.M Alimi
Research Group on Intelligent Machines
Engineering School of Sfax (ENIS), University of Sfax
BP 1173, Sfax, 3038, Tunisia

## ABSTRACT
This paper describes a new approach allowing the generation of a simplified Biped gait. This approach combines a classical dynamic modeling with an inverse kinematics' solver based on particle swarm optimization, PSO. First, an inverted pendulum, IP, is used to obtain a simplified dynamic model of the robot and to compute the target position of a key point in biped locomotion, the Centre Of Mass, COM. The proposed algorithm, called IK-PSO, Inverse Kinematics PSO, returns and inverse kinematics solution corresponding to that COM respecting the joints constraints. In This paper the inertia weight PSO variant is used to generate a possible solution according to the stability based fitness function and a set of joints motions constraints. The method is applied with success to a leg motion generation. Since based on a pre-calculated COM, that satisfied the biped stability, the proposal allowed also to plan a walk with application on a small size biped robot.

## Keywords
Biped robotics, Gait generation, Particle Swarm Optimization, Inverse kinematics. Inertia weight PSO.

## 1. INTRODUCTION
A legged system is a robotic system that has a specific mission, ensuring locomotion. Such a system could be analyzed from its kinematics aspects and also its dynamics aspects. The kinematics aspects are centered on the joints motions, constraints and limits. Forward kinematics is the mathematical relationship between the end segments of an articulated system and its joints motions. In inverse kinematics, the problem consists of computing the needed joints motion s leading a known end segment position [1].

Biped robots are specific robotic system that combines open and close kinematics chains; in biped locomotion two key phases are important, the double support phase and the single support phase also known as stance phase and swing phase. In double support both legs are in contact with the flow the obtained articulated body tend to move its COM, Centre of mass, position to prepare the next step, in single support or swing phase a leg is used as a support while the opposite one swings forwards, this motion causes the displacement of the COM in the same direction of the swing, when the swing leg is once again in contact with the floor the COM is placed on its surface and that gives a forward step.

The forward kinematics model of a robot could be simply obtained using the classical Denavit-Hartenberg method [2], while the inverse kinematics problem is little bit more complex. A classical approach to inverse kinematics consist in in writing the inverse mathematical formulation of the forward model, This approach is know to be time and memory consuming. The classical inverse modeling process based on geometric [3], algebraic or iterative methods is limited if the system structure is complex [4].

Many alternative methods are proposed to solve inverse kinematics problem. To tackle the inverse kinematics modeling an ANFIS (adaptive neuro fuzzy inference system) methodology were introduced in [5]. Basically, ANFIS methods used a set of data representing the end points of an articulated system and the joints motions needed to produce them to generate neuro-fuzzy network able to solve inverse kinematics. Genetic and evolutionary algorithms were also investigated as possible solutions to complex inverse kinematics problems [6] [7], combination of genetic and Particle swarm optimization was also investigated [8].

In this work the constrain weight particle swarm algorithm is investigated to solve inverse kinematics problem with a specific focus on biped robotics, including leg motion and also walking gaits.

The paper is organized as follows; section2 introduced the robotic modeling with both aspects kinematics and dynamics before detailing given a brief on particle swarm optimization. Section 3, is dedicated to IK-PSO, inverse kinematics PSO with application to swing leg motion generation and walking gaits of a biped robot. Experiments and results are detailed in section 4. The paper ends by a discussion the results and further works, in section5.

## 2. MODELING IN BIPED ROBOTICS
A simplified biped robot could be observed as an articulated system with a specific configuration allowing it to be an open loop kinematics system or a closed loop kinematics one. A simplified 2D representation on the sagittal frame is showed in figure 1 (a); its equivalent inverted pendulum appears in figure1 (b).

For a planar robot, the kinematics modeling process consists in finding a solution to the end segments of each leg, (hip, knee and ankle) according to joints motions. Joints are supposed to produce rotations the hip joint motion is responsible of the knee position, the knee joints is in charge with the ankle end position and finally the ankle joint is responsible of the foot position.

For kinematic-modeling needs a set of frames are used, a key frame is the reference frame, for biped robots this frame is used to compute a specific point called, COM, centre of mass. The COM is a virtual point representing the centre of gravity of the body according to the relative position of joints and masses see figure 1 (b). A simplified dynamic model of a biped robot could be obtained using the inverted pendulum made by the COM and a virtual support leg that joined the COM to the ground. The IP model allowed an estimation of the needed torque to support the overall robot structure, see paragraph 2.2, dynamic modeling.

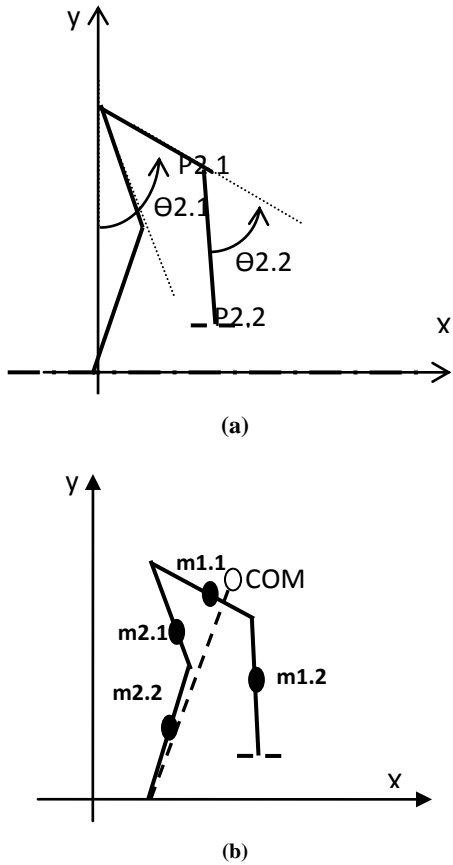

**Figure 1. Simplified biped robot, (a) segments based representation, (b) equivalent inverted pendulum estimation with a COM of the locomotion system.**

An important issue in biped robotics is the stability of the structure while walking [9]. Any motion has a direct impact on the stability and could lead to a fall down. Handling the kinematics of such a system should include both, the motion generation it self and the impact of the motion on the structure stability. The COM helps also a simple control of the stability; a biped is stable since its COM projection on the floor remains within the sustention polygon, with is the surface of foot prints wit respect to the walking cycle [10].

## 2.1 Kinematics model of an articulated system

Assuming an articulated system composed by a set of links and joints, to build a kinematics model we need a set of frames placed in specific points, generality the links end segments. For a robotic system with (n) end-segment positions, called $P_i = (P_1, P_2, ..... P_n)_i$ , a set of joints rotations correspondent to them respectively $q_i = (\theta_1, \theta_2, ...... \theta m)_i$ , and a set of translations Ti, where (i) stands for the iteration number, so the forward kinematics could be expressed by equation (1).

$$Pi = f(qi, T_i) \qquad (1)$$

Classically the forward model is build using a set of geometric transformations; it could be obtained using a simplified approach such us the Denavit-Hartenberg one [11]. In this case the robot is observed as a set of links and joints, some rules should be observed when placing the frames on the links so that it will become easy to obtain the forward kinematics model.

The inverse kinematics is a function that gives, when it is defined, a set of angular rotations of the joints and elementary translations Ti in order to achieve a given position St. that could be expressed by the equation (2)

$$(Q_i, T_i) = f^{-1}(Pi) \qquad (2)$$

If the system joints motions are limited to rotations the inverse kinematics problem is then establish as in equation (3).

$$Q_i = (q_1, q_2, ..... q_n) = f^{-1}(Pi) \qquad (3)$$

The $\dfrac{\partial f}{\partial q_i} = J$ is known to be the Jacobian's of f() witch is no more that the first order derivative of f(). If J is invertible, the inverse kinematics is easy to compute using equation (3), if not we have to approximate J and to use the approximation to obtain a solution.

$$\Delta Q_i = J^{-1} * \Delta P_i \qquad (4)$$

Several methods are proposed, the Jacobian transpose methods uses transpose of J instead if its inverse, the pseudo inverse method replaces the Jacobian by its pseudo inverse, see equation (4), while it is not the exact inverse of J it still a good approximate on of it [4] .

## 2.2 Dynamic modeling of a biped robot

The inverted pendulum, **IP**, gives an approximation of the biped robot dynamic model; it allowed also the estimation of the robot stability [9]. In this case the IP supports a specific mass called the COM witch is virtual mass with a virtual position P(x,y,z); its linear velocity is assumed to be the linear velocity of the robot, and its acceleration the linear acceleration of the hole body too. The fundamental equation of dynamics could then be written as in (6). It gives an approximation of the torque needed in the angle joint of the support leg.

To establish a simplified model, the leg is represented by a set of masses and segments see figure 1 (a), where segments sizes and masses depends on the robot mechanical design. Known the legs segments masses and relative positions, the COM position, **P (Xcom, Ycom)**, could be computed using equations (5)[10].

$$x_{com} = \frac{1}{\sum_{i=1}^{n} m_i} \sum_{i=1}^{n} x_{mi} * mi$$

$$y_{com} = \frac{1}{\sum_{i=1}^{n} m_i} \sum_{i=1}^{n} y_{mi} * mi \qquad (5)$$

$$z_{com} = \frac{1}{\sum_{i=1}^{n} m_i} \sum_{i=1}^{n} z_{mi} * mi$$

*P (x,y,z)*, is the position vector in a 3D (x,y,z) frame. The planar representation used here is limited to (x,y). Considering the relative IP, leg angular position , θ, and τ is the torque needed to ensure the COM placement.

$$m\|\vec{g}\|*\cos(\theta) + m*\|\vec{a}\|*l*\sin(\theta) = \tau$$

$$\vec{a} = \frac{\partial^2 (x_{com})}{\partial t^2}*\vec{i} + \frac{\partial^2 (y_{com})}{\partial t^2}*\vec{j}$$

(6)

Equation (5), represents also the equation of motion of the COM, allowing a pres-computed COM target position, this position target position is then used to compute the necessary joints motions

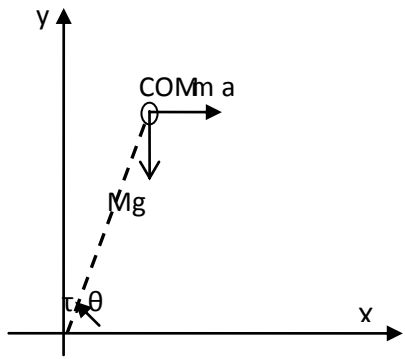

**Fig 2: Inverted pendulum, a simplified representation of a biped robot in single support.**

## 2.3 Particle swam optimization

Particle swarm optimization, PSO, was first investigated by Eberhart and Kennedy as a computational approach to solve complex problems simply by mimic the social comportment of animals. In a group each individual is represented by a particle with a position and a target mission [12].

In PSO, a problem is addressed threw a search space and a fitness function; a set of particles are randomly spread on the search space, and asked to find a solution, each particle is a potential solution of the problem, while the best one is the one fitting the fitness function. The swarm is generally asked to iterate until the fitness function is satisfied or to the end of a fixed iteration number. For each iteration the particles positions on the search space are updated according to a couple of equations see equation (7) and (8). Those equations are also used to formalize the interactions that could occurs within the group, this mimic the bird folks or the fish banks global positioning in witch the group has a comment objective, individual comportments still allowed within the group global policy and has a direct impact on it. The native PSO formulation appears in equations (7) and (8).

$$\vec{V}_{i+1} = \vec{V}_i + c_1*rand()*(\vec{x}_{lbest} - \vec{x}_i) + c_2*rand()*(\vec{x}_{Gbest} - \vec{x}_i)$$ (7)

$$\vec{x}_{i+1} = \vec{x}_i + \vec{V}_{i+1}$$ (8)

Where *(V)* stands for the velocity, *(c1)* and *(c2)* are called acceleration factors and *(x)* is the particle position. The velocity is a displacement of the particle, according to its best local and to the best particle of the swarm. Setting parameter

c1 and c2 as equal will give equivalent potential weight to the local search strategy and the global search one. The local best is elected between particle neighborhoods; a neighborhood topology should be chosen before running PSO.

A key issue in PSO is to establish a particle representation that fits the problem formulation and to define a fitness function that ends the solving process. It is also important to remind that PSO gives always a solution, even if the fitness function is not totally satisfied, the best particle is returned as a solution within the iteration number limit.

According to [12], the PSO algorithm runs as follows:

1) Initialize a population of particles

2) For each particle evaluate the fitness function

3) Compare fitness function to local best particle one, and if better set the particle as local best.

4) Compare fitness to global best particle, and if better set new global best position as current particle position.

5) Update the particles positions according to equations (4) and (5).

6) Loop to step (2) until stop criterion is met, (**fitness function satisfied or maximum iteration number achieved).**

It is also common to control and limit the velocity within a know range by fixing the maximum amount that the velocity could achieve, **Vmax.** Several variations were then introduced in order to optimize the native proposal. Inertia weight PSO is one of the major variants of PSO [13], it allowed to moderate the velocity using an inertia coefficient allowing to smooth or to fasten the search procedure.

On the other hand PSO was investigated as an optimization technique used in conjunction with other intelligent methods such as neural networks, genetic algorithms or ant systems [14].

## 3. IK-PSO, INVERSE KINEMATICS USING PSO

Particle swarm optimization was proposed for motion generation and optimization; attempts to use original PSO formulation to generate the gaits of biped robotic system [15] [16] [17]. In [15] a proposal using the symmetric characteristics' of the biped walking gaits to run a PSO that transfers the legs motions characteristics from one to another. In [16] a random gait generator using classical PSO [11] with a walking stability control was investigated, the proposal was then tuned with agent based architecture [17]. In [18] a PSO gait generator where inserted into a control flow using classical PID controllers. In this work the PSO is investigated as an alternative to IK solvers. The proposed algorithm could be applied to any robotic system, while in this paper the focus is made only on its use for biped robot.

## 3.1 IK-PSO algorithm

The proposed IK-PSO solver, could be resumed as follows, first we assume that the particle is a vector of joint rotations Qi, so swarm will work in the joints frames and a particle represents a potential solution of the articulated system. Then the forward kinematics model corresponding to that solution is used to establish the resultant motion of the links. A specific point is considered as a key point for the motion optimality, it could be an end-segment, or a virtual point such the COM of a biped robot.

The following PSO formulation is proposed in [18], it has the advantage to fit better the requirement of our problem. PSO equations are introduced in (9) and (10)

$$\vec{V}_{i+1} = w * \vec{V}_i + \varphi_1 * (\vec{q}_i^{lbest} - \vec{q}_i) + \varphi_2 * (\vec{q}_{best} - \vec{q}_i) \quad (9)$$

$$\vec{q}_{i+1} = \vec{q}_i + \vec{V}_{i+1} \quad (10)$$

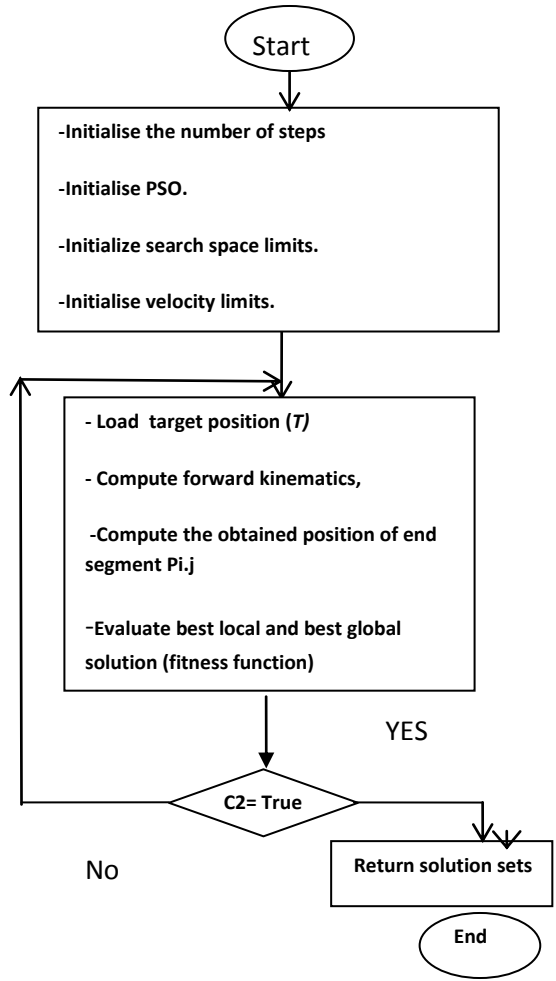

**Fig 3: Simple chart of forwad walk using a limited number of steps, C2 is the end condition with is the end of steps.**

## 3.2 Swing leg problem formulation

To apply the IK-PSO algorithm on simple double segment leg model, we have first to write the forward kinematics of that system. The forward kinematics of a leg 1, as it appears in figure 4, included two joints and two links. For each leg (i), **Pi.1** and **Pi.2** represent respectively the end-segments positions of link (i.1) and link (i.2). Where link (i.1) refers to the first link of leg (i), the opposite leg is here the support leg and is considered as a fix leg. For each leg the particle representations are in (11)

$$Q = (q_1, q_2) = \left[ (\theta_{1.1}, \theta_{1.2}), (\theta_{2.1}, \theta_{2.2}) \right]$$

$$under \begin{cases} \theta_{i.1} \in \left[ \theta_{1\min}, \theta_{1\max} \right] \\ \theta_{i.2} \in \left[ \theta_{2\min}, \theta_{2\max} \right] \end{cases} \quad (11)$$

The forward model of leg (i) is expressed by equation (12).

$$\begin{cases} Xi.1 = l1 * \cos(\theta 1.1) \\ Yi.1 = l1 * \sin(\theta 1.1) \\ Xi.2 = l1 * \cos(\theta 1.1) + l2 \cos(\theta 1.1 + \theta 1.2) \\ Yi.2 = l1 * \sin(\theta 1.1) + l2 \sin(\theta 1.1 + \theta 1.2) \end{cases} \quad (12)$$

Where (*l1*) is the length of link1.1 and (*l2*) is the length of link 1.2. Since the legs are symmetric, (l1) and (l2) are respectively the same for leg 2. Equation (12) expresses the kinematic model in the frame (X1, Y1), see figure 4.

Equation (11), gives the forward model of the leg i, for both legs a simplified model is used, and the foot print is neglected. For a swing motion, only one leg is moving while the opposite one is assumed to be the support one.

The inverse kinematics problem of this swing leg could be written as follows: for a target position of P2.2, T2.2; find the corresponding joints positions. Additional constraint could be also added. The fitness function here could be simply the distance of P2.2 to the T2.2, see figure4.

$$fitenss_i = \left\| T2.2 - P2.2 \right\|^2 \quad (13)$$

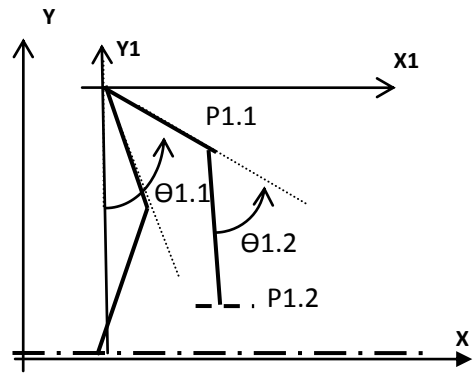

**Fig 4: Simplified biped robot, segments based representation, the dashed line represent the floor level**

## 3.3 Walking gait problem formulation

For walking gait generation, the same particles representation is used as in equation (11); the sagittal forward kinematics model of the system is written bellow, equation (13). The simplified equation of motion of the COM, detailed in paragraph 2.2, is used to generate a COM reference trajectory, and then IKPSO is then used to obtain the leg gaits. Since legs have symmetric comportment, only the swing leg is represented, the support leg is assumed to fix, the problem formulation is easier.

$$\begin{cases} X1.1 = l1*\cos(\theta1.1) \\ Y1.1 = l1*\sin(\theta1.1) \\ X1.2 = l1*\cos(\theta1.1) + l2\cos(\theta1.1 + \theta1.2) \\ Y1.2 = l1*\sin(\theta1.1) + l2\sin(\theta1.1 + \theta1.2) \\ X2.1 = l1*\cos(\theta2.1) \\ Y2.1 = l1*\sin(\theta2.1) \\ X2.2 = l1*\cos(\theta2.1) + l2\cos(\theta2.1 + \theta2.2) \\ Y2.2 = l1*\sin(\theta2.1) + l2\sin(\theta2.1 + \theta2.2) \end{cases} \quad (13)$$

The particles representation and the fitness function for case appears in equations (14) and (15).

$$q = \left[ (\theta_{1.1}, \theta_{1.2}, \theta_{2.1}, \theta_{2.2}) \right] \quad (14)$$

$$fitenss_i = \| ComT - Com \|^2 \quad (15)$$

The gait generation method could be resumed as follows :

1) Compute COM reference trajectory (comT).

2) Initialize a population of particles

3) For each particle evaluate the fitness function

4) Compute the forward kinematics model

5) Compute obtained COM position (com).

6) Compare fitness function to local best particle one, and if better set the particle as local best.

7) Compare fitness to global best particle, and if better set new global best position as current particle position.

8) Update the particles positions according to equations (4) and (5).

9) Loop to step (2) until stop criterion is met, (*fitness function satisfied or maximum iteration number achieved*).

## 4. EXPERIMENTAL RESULTS

### 4.1 IKPSO for swing leg motion generation

To test the IK-PSO algorithm a target position was fixed for the swing leg, the algorithm was run with a stop condition witch is fitness satisfaction or the end of iteration numbers, the maximum number of iteration was fixed to 5000, and witch is a commonly used parameter in standard PSO processing.

Fig 5, illustrates the solutions set of all possible motions that could be possible for a double segment articulated system, here no constraints were imposed on the search space. For a known target point, a solution appears in red, while not unique, it is the one with the best fitness function. For this case no limits or specific constraints were applied to joints angles, the system was left free to find any solution, and the fitness is simply the minimum distance between the end segment and the target

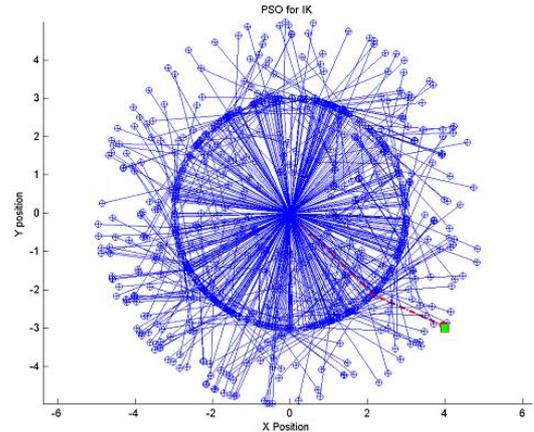

**Fig 5: PSO solving a simple IK problem, here a robot articulated system composed of two segments, for a target point that appears in green, the best solution is in red.**

At the opposite of the previous example where we was interested only by the target position and particles velocity, figure 6, illustrates a swing leg motion in witch search spaces were limited for both joints with constraints as in equation (12). For both experiments the fitness function target appears in (11) and here and end-segment target position was used.

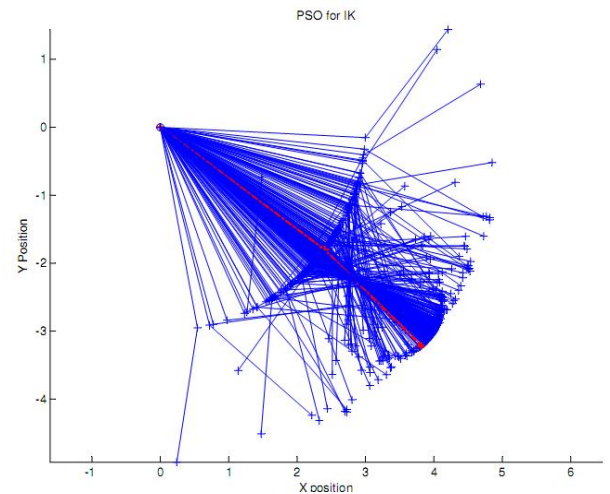

**Fig 6: Simplified leg gait generation using PSO**

### 4.2 Robot Prototyping

A robotic Kit is used here to validate the IK-PSO method [19]. For that, the assembled robot has a design close to the proposed simplified model, see figure 7. In this assembled biped, a leg is composed by a two segments and a footprint. Joints are actuated using servomotor that includes its controller with a build in PID controller.

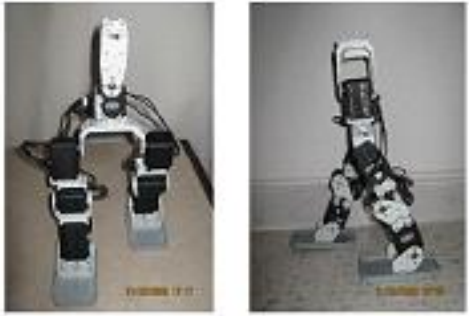

**Figure 7. biped robot prototype**

The walking gait is obtained by generating a swing motion of leg 1, while leg 2 is use to support the robot, and a swing motion of leg 2 when leg one is the support one.

In our case the following steps were followed during the robot prototyping:

- First we build a simple biped robot with only three degree of freedom on each leg.

- Then the robot parameters were measured including, segments sizes, servos masses.

- Real robot motion angular constraints were report to the IK-PSO algorithms

- Using the inverted pendulum simplified model the COM target position was approximated.

- Distance of generated COM to targeted Com was used as a fitness function of the IK-PSO.

The detailed prototyping methodology is detailed in [19]. And the experiment leads to a successful walk of the robot.

## 5. DISCUSSION AND FURTHER WORKS

In this paper we introduce a new proposal for motion generation of a biped robotic system. The proposal is based on an inverse kinematics solver that uses inertia weight particle swarm optimizer, PSO. The IK-PSO algorithm, Inverse Kinematics PSO, is based on a forward kinematic model, a specific formulation of PSO and a fitness function to give the inverse kinematics solution of a robotics system.

The IK-PSO generates a set of solutions, each one in a particle, then compute the forward equivalent target position of a specific point, the COM for a walking gait, and returns the best one.

The algorithm could solve any inverse kinematics problem, while here applied to the swing motion leg of a biped robot before being extended to the global walking gait, where a simplified inverted pendulum biped model is used to estimate the COM position. This position was then given to the IK-PSO solver in order to generate the correspondent angular positions necessary for the joints motion.

For validation needs a robotics educational kit were used to build a simple biped, its parameters, were then injected to the IK-PSO in order to generate the joints rotations and control the robot walking with success.

## 6. ACKNOWLEDGMENTS

The authors would like to acknowledge the financial support of this work by grants from General Direction of Scientific Research (DGRST), Tunisia, under the ARUB program.